# Enhancing underwater image via adaptive color and contrast enhancement, and denoising

Xinjie Li, Guojia Hou, *Member, IEEE*, Kunqian Li, *Member, IEEE*, and Zhenkuan Pan

*Abstract*—Images captured underwater are often characterized by low contrast, color distortion, and noise. To address these visual degradations, we propose a novel scheme by constructing an adaptive color and contrast enhancement, and denoising (ACCE-D) framework for underwater image enhancement. In the proposed framework, Difference of Gaussian (DoG) filter and bilateral filter are respectively employed to decompose the high-frequency and low-frequency components. Benefited from this separation, we utilize soft-thresholding operation to suppress the noise in the high-frequency component. Specially, the low-frequency component is enhanced by using an adaptive color and contrast enhancement (ACCE) strategy. The proposed ACCE is an adaptive variational framework implemented in the HSI color space, which integrates data term and regularized term, as well as introduces Gaussian weight and Heaviside function to avoid over-enhancement and oversaturation. Moreover, we derive a numerical solution for ACCE, and adopt a pyramid-based strategy to accelerate the solving procedure. Experimental results demonstrate that our strategy is effective in color correction, visibility improvement, and detail revealing. Comparison with state-of-the-art techniques also validate the superiority of proposed method. Furthermore, we have verified the utility of our proposed ACCE-D for enhancing other types of degraded scenes, including foggy scene, sandstorm scene and low-light scene.

## I. INTRODUCTION

As more and more scarce resources have been found in lake, river and ocean, underwater imaging receives growing attentions in recent years. Unlike the common outdoor images, underwater scenes always suffer from a poor visibility because of color cast and contrast loss. In-depth researches [1], [2] have revealed that this problem is caused by the propagated attenuation when light travels through water, primarily due to absorption and scattering. The absorption reduces the light energy, while the scattering forces a change in the direction of light travel. The color distortion is caused by light of different wavelengths decaying in water at different rates. For example, the red component, which has the longest wavelength, rapidly vanishes in about 3 meters distance, while the shorter green and blue components reach a greater depth, and gradually dominate the main tones of underwater image. As a result, images captured under water are visually unpleasing, and can hardly be directly used to carry out further studies, as shown in Fig. 1.

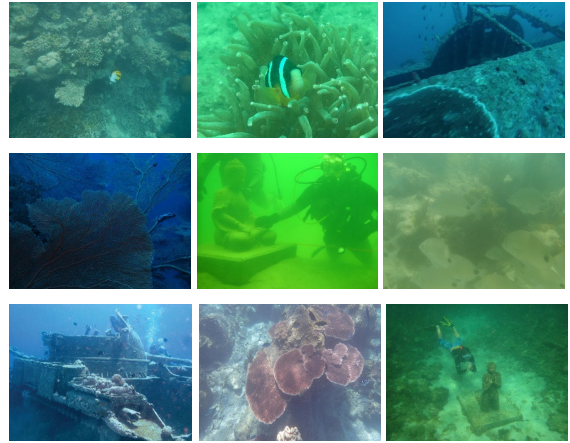

Fig. 1. Examples of underwater images with different degradations.

To address these problems, a series of underwater image restoration and enhancement methods have been proposed to improve the quality of underwater images. According to the treating processes, the existing methods can be generally divided into three categories: image restoration-based methods, image enhancement-based methods and deep learning-based methods. Image restoration-based methods [3]-[5] utilize the underwater image formation model [6], [7] and invert it to acquire the non-degraded image. On the contrary, image enhancement-based methods [8]-[10] utilize the effective image processing techniques to enhance some characteristics of image, aiming to produce visual pleasing results. More recently, deep learning has received extensive attention because of its remarkable performance on image processing. Several deep learning-based methods [11]-[13] are also specially explored to improve the quality of underwater image.

Despite some breakthroughs made in recent years, it is still challenging to address the light absorption and scattering problems. Actually, some exiting methods cannot comprehensively address the multiple degradation problems existing in underwater images improvement. In some examples, the enhanced or resorted results may be accompanied with inadequate contrast, inaccurate color correction, minimal brightness improvement and noise amplification. Specially, the

This work was supported in part by the National Natural Science Foundation of China under Grant 61901240, in part by the Natural Science Foundation of Shandong Province, China, under Grant ZR2019BF042, in part by the China Scholarship Council under Grant 201908370002, and in part by the China Postdoctoral Science Foundation under Grant 2017M612204. (Corresponding authors: Guojia Hou).

Xinjie Li, Guojia Hou and Zhenkuan Pan are with the College of Computer Science and Technology, Qingdao University, Qingdao 266071, China (email: xinjie_li68@163.com, hgjouc@126.com, zkpan@126.com).

Kunqian Li is with College of Engineering, Ocean University of China, Qingdao 266100, China (likunqian@ouc.edu.cn).

majority problem of enhancement-based methods exhibits to overstretch the contrast, which may lead to the loss of valuable information. Therefore, it is expected to develop an effective method overcome these shortcomings.

In this paper, we propose a novel strategy for underwater image enhancement by adaptively enhancing color and contrast, and suppressing noise. Our key contributions and works are summarized as follows:

(1) Taking advantage of noise typically distributes in the high frequency component, we develop a decomposition scheme to avoid noise amplification in the low frequency component. The proposed scheme can not only reduce the noise effect, but also sharpen the important edge blurred by the forward-scattering.

(2) A new adaptive variational framework ACCE is formulated for color and contrast enhancement, in which we exploit the Heaviside function to divide the image channel into two subsets, and adopt two different enhancement strategies for each subset to avoid the over-enhancement and oversaturation.

(3) In order to reduce the computational complexity of proposed framework, a numerical approximation combining the gradient descent and a pyramid-based acceleration strategy is designed to optimize the whole progress.

(4) The proposed method doesn't require any prior knowledge or optical process, so it is more robust and can be generalized to improve the quality of other types of degraded images. Extensive experiments demonstrate that our scheme is effective in color correction, contrast enhancement, and sharpness promotion.

The rest of this paper is organized as follows. In section II, we introduce the related work by presenting a brief overview of the existing state-of-the-art methods of improving underwater image quality. In section III, the proposed ACCE-D method is presented and analyzed. Section IV is devoted to discuss the experimental results as well as comparison to advanced techniques. Finally, we summarize our work in section V.

## II. RELATED WORK

In this section, we give an overview of existing related work of improving underwater images involving restoration–based methods, enhancement-based methods, and deep learning-based methods.

As mentioned above, image restoration-based methods rely on the underwater image formation model (UIFM) [7], [8] and physical assumption to dehaze or deblur the degraded scene, as to recovery the desired results. Since UIFM has a close form to the hazy image formation model proposed by Koschmieder [14], which characterizes the propagation progress of light in atmosphere. Therefore, some method designed for outdoor scene dehazing can also be applied underwater after a reasonably modification. The famous dark channel prior (DCP) method [15] assumes that the haze-free image has a very low value in at least one channel. Based on this hypothesis, He et al. inversed the hazy imaging model to estimate the restored results. Drews, Jr., et al. [16] thought that for the underwater environment, the red component can not provide sufficient visual information since it decays too fast. They hence calculated the dark channel image by only using the green and blue color channels. On the contrary, Galdran et al. [17] utilized the reciprocal of red component to estimate the scene depth, and proposed a novel red channel prior (RCP). Peng et al. [18] incorporated the adaptive color correction into the IFM and developed a general restoration method on the basis of DCP. Beyond these, several methodologies [19]-[21] were proposed to improve the accuracy of transmission map estimation. However, since it is unrealistic that single assumption or water properties are applicable to various challenge underwater environments, this kind of methods only perform well in some given scenes. In addition, these methods may have no advantages in terms of color correction.

Different from the restoration-based methods, image enhancement-based methods aim to generate visually pleasing results by directly modifying image pixel values. Some traditional enhancement methods (e.g., histogram equalization [22], [23], Gray-world [24], [25]) have been applied for improving the underwater image quality. Despite these methods can improve the image quality in some aspects, they are seldom used alone due to the neglecting of other factors that may cause the visual degradation. Iqbal et al. [26] stretched the dynamical range of pixels in RGB and HSI space, aiming to equalize the color and contrast. In [27], Ghani and Isa modified the schemes of [26] and proposed a series of stretching strategies based on the Rayleigh distribution. Experimental results show that their methods are able to ameliorate the visibility meanwhile reduce the over-enhancement or under-enhancement regions. Ancuti et al. [28] applied multi-scale image fusion technology for underwater dehazing task. They used a single image to generate one color corrected version and another contrast enhanced version, then fused them with defined weights to obtain the desired result. More recently, this fusion strategy was further improved in [29] to cope with more severe imaging environment. Marques et al. [30] combined the local contrast and fusion techniques to specifically enhance the underwater low-light images. Fu et al. [31] utilized a two-step scheme including color correction and optimal contrast improvement algorithms to address the problem of color distortion and low contrast. Later, in [32], they propose a variational framework to enhance single underwater image based on the Retinex theory. Zhang and Wang [33] extended the muti-scale retinex (MSR) to LAB-MSR for improving the quality underwater image. Compared with the traditional MSR, their method not only increases the visual quality, but also mitigates the halo-artifact. Afterward, Zhang et al. [34] developed a new method using color correction and Bi-interval contrast enhancement to enhance underwater image.

Different from restoration-based and enhancement-based methods, deep learning technology utilize the artificial neural network to automatically extract features and learn mapping functions from training set. For the case of underwater image, many deep learning-based methods have been proposed to estimate the non-degraded scene. Li et al. [35] developed a weakly supervised color transfer method for the color correction. In [36], a multiscale dense generative adversarial network (GAN) is presented for underwater image enhancement. Chen et al. [37] employed a GAN-based restoration scheme to adaptively improve underwater image quality in real time. Islam et al. [38] also proposed a GAN-based approach to enhance underwater image and constructed a large-scale dataset composed of a paired and unpaired set of underwater images. In [39], Li et al. proposed an underwater scene prior and developed a convolutional neural network for enhancing underwater image. Fu et al. [40] designed a novel deep learning method based on global-local network and compressed-histogram equalization (GNCE) to solve the visual degradation in complex and changeable underwater scenes. Yang et al. [41] utilized a conditional generative adversarial network with dual discriminator to improve visual quality of underwater image. Besides, a real-world benchmark dataset is built in [42] and used for training an end-to-end underwater image enhancement network. Qi et al. [43] proposed a novel Co-Enhancement network for enhancing underwater image based on an encoder-decoder Siamese architecture. Despite achieving notable progress, deep learning-based methods highly depend on the data-driven. In addition, due to the complicate underwater condition, it is very difficult to collect effective and sufficient real-world and synthetic underwater images for training deep networks.

In contrast to existing algorithms, the proposed method doesn't directly enhance the contrast of image, but manipulates in where we called low-frequency components, so that the noise effect can be isolated. Compared with existing contrast-enhancement techniques, our method is more effective in local and global contrast boosting, as well as produces less information loss since the enhancement strategy is adaptively adjusted according to the local conditions. Therefore, our ACCE-D method can significantly improve the image quality in terms of contrast, color, and brightness. Moreover, as our proposed method doesn't rely on any physical model, it is also robust and can be well generalized to other degraded scenes.

## III. PROPOSED METHOD

Sourced from Retinex theory [44], the automatic color equalization (ACE) [45], [46] is an effective method for color and contrast enhancement. By considering the spatial, local and nonlinear features of the human visual system (HVS), ACE can generate more natural results, which is well consistent with perception. In [47], Bertalmío et al. generalized it to a variational form, which allow us to extract the desired characteristics from ACE. The energy functional in [47] is established as:

$$E(I) = \frac{\alpha}{2} \sum_{x \in \Omega} \left( I(x) - \frac{1}{2} \right)^2 + \frac{\beta}{2} \sum_{x \in \Omega} \left( I(x) - I^0(x) \right)^2 - \frac{\gamma}{2} \sum_x \sum_y w(x,y) S \left( I(x) - I(y) \right) \tag{1}$$

where $I$ and $I^0$ denote one channel of enhanced result and original image, respectively. $x$, $y$ are the pixel coordinates on image plane. $w$ is the distance weighted function, its value decreases along with the increase of the distance between $x$ and $y$. $S$ is the primitive of the slope function that accounts for the relative intensity difference. By performing the minimization of (1), an enhanced result can be produced. However, when comes to the more challenging underwater scenes, ACE often fails to achieve the visual appealing results due to its inaccurate color correction strategy. To illustrate this, we have exhibited several enhanced images generated by ACE algorithm as shown in Fig. 2. Besides, the enhanced results by using our proposed method are also accordingly presented. From Fig. 2, we can observe that ACE fails to correct the color cast precisely. But even so, the contrast enhancement of ACE is still meaningful for us because the local-global mechanism is contained.

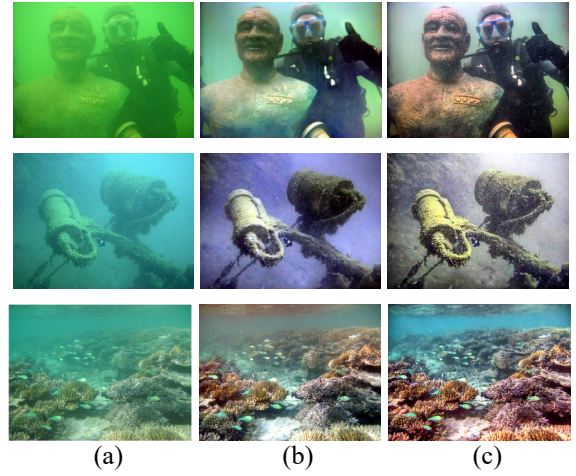

Fig. 2. Comparisons of the results of ACE and proposed method. (a) Raw images, (b) the enhanced results of ACE method, (c) the enhanced results of proposed method.

Inspired by ACE, we propose a novel enhancement strategy ACCE-D to improve the visibility of underwater image without relying on any optical model inversion or physical assumption. Since the main information of an image is mainly contained in the low frequency component, and the edge and texture information are mainly contained in the high frequency component. Therefore, we first decompose the image into the low-frequency component and the high-frequency component. In our framework, we adopt the bilateral filter [48] to obtain low-frequency information for extracting more valuable information. For removing the color deviation as well enhancing contrast, we construct an adaptive variational framework for underwater image enhancement, dubbed ACCE. Since the hue channel H, saturation channel S, and intensity channel I are independent of each other in the HSI color space. We convert it from RGB color space to the HSI color space, in which we perform the ACCE operation to generate an enhanced result by color correction and contrast enhancement. Moreover, we design a fast numerical solution combing with the gradient descent and a pyramid-based acceleration strategy for solving the proposed ACCE framework. For the high-frequency component, we employ the Difference of Gaussian (DoG) [49] filter to separate it from the original image, and subsequently perform a simple soft thresholding operation [50] to reduce the noise to avoid mistaking it for an edge. Finally, we utilize a simple linear combination to integrate the enhanced low-frequency component and the denoised high-frequency component. The workflow of our method is presented in Fig. 3.

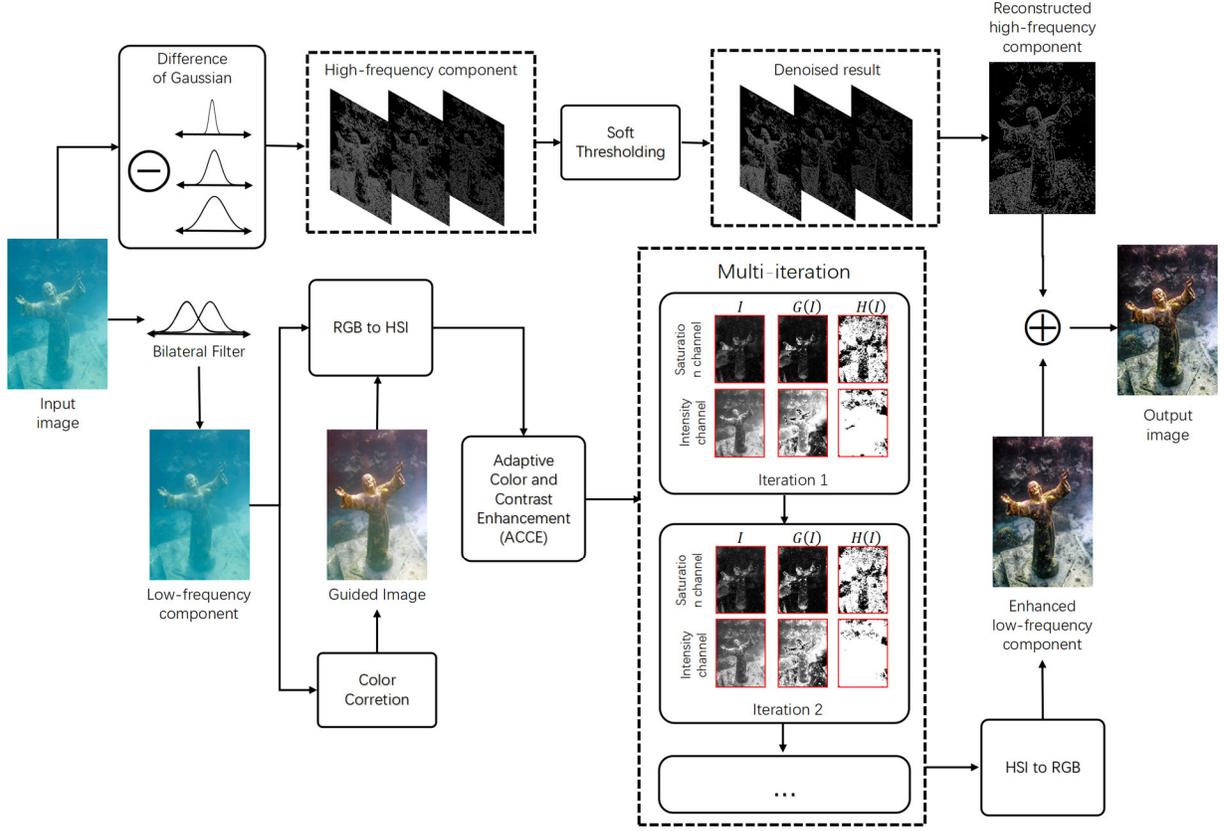

Fig. 3. Flowchart of the proposed ACCE-D method.

As the core of our scheme, in the following, we will give the detailed explanation of the proposed ACCE framework. Denoting that all inputs are normalized to the interval [0, 1], a color and contrast enhancement framework is initially established as:

$$
\begin{cases}
E(I) = \dfrac{1}{2} \sum_{\theta \in h, s, i} \sum_{x} \left( Cr_{\theta}(x) - I_{\theta}(x) \right)^2 \\
\qquad - \dfrac{\alpha}{2} \sum_{x} \sum_{y} w(x, y) \left( I_s(x) - I_s(y) \right)^2 \\
\qquad - \dfrac{\beta}{2} \sum_{x} \sum_{y} w(x, y) \left( I_i(x) - I_i(y) \right)^2, \\
s.t. \begin{cases} I_s(x) = 0 & I_s(x) < 0 \\ I_s(x) = 1 & I_s(x) > 1 \end{cases}, \\
s.t. \begin{cases} I_i(x) = 0 & I_i(x) < 0 \\ I_i(x) = 1 & I_i(x) > 1 \end{cases}
\end{cases}
\tag{2}
$$

where $I$ is the enhanced result. $x$ and $y$ represent two pixels on the image plane. $h$, $s$, $i$ represent the three channels in HSI color space. $Cr$ denotes a guided image with color correction. The constrains in (2) are used to prevent the values of $I_s$ and $I_i$ from exceeding the range [0, 1].

To generate the $Cr$, we employ a simple yet effective operation [32] by calculating the mean values and the mean square error (MSE) on each channel in RGB space. Then, the upper and lower threshold of each channel can be obtained by:

$$
\begin{cases}
UT_{\chi} = avg(f_{\chi}) + \lambda \cdot mse(f_{\chi}) \\
LT_{\chi} = avg(f_{\chi}) - \lambda \cdot mse(f_{\chi})
\end{cases}
\tag{3}
$$

where $f$ is the low-frequency component of input image, $UT$ and $LT$ are the upper and lower thresholds. $\chi \in \{R, G, B\}$ denotes the color channel, $\lambda$ is a parameter related to the image dynamic, which is often empirically set to 2.3. Subsequently, the color corrected image can be obtained by stretching the histogram of the pixels within the upper and lower thresholds.

$$
Cr_{\chi}(x) = \frac{f_{\chi}(x) - LT_{\chi}}{UT_{\chi} - LT_{\chi}} \times 255
\tag{4}
$$

It can be easily observed from (2) that the proposed variational model constructs a competitive relationship between one positive term and two negative terms. The positive term is utilized to correct the trend of color variance by penalizing the deviation of $I$ from the $Cr$. However, the other two negative terms attempt to amplify the difference of the weight $w$ between each pixel in S and I channels, which is beneficial to local and global contrast enhancement. By minimizing (2), we can acquire an enhanced outcome characterized by high contrast and bright color. But unfortunately, some pixels (saturation or brightness values are close to 0 or 1) will lie outside of the range [0, 1] because of excessive enhancement. This maybe lead to produce some brighter or darker areas in the enhanced result. More seriously, some valuable information may also be lost. Fig. 4 displays the enhanced results yielded by minimizing (2). It can be easily seen that the results generated from (2) are not satisfactory due to over-enhancement.

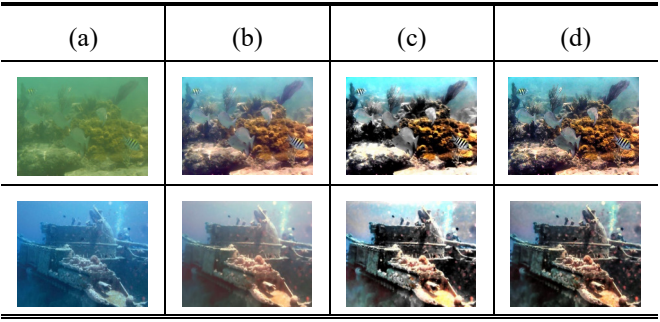

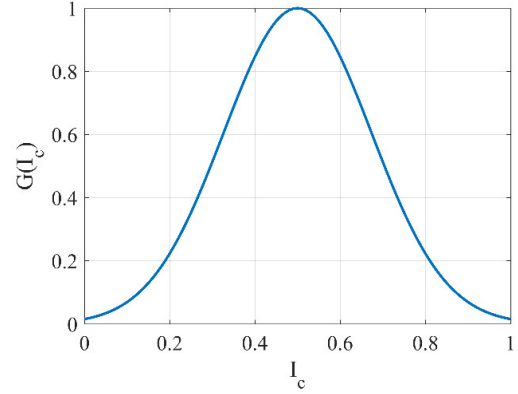

Fig. 4. Example of underwater image enhancement. (a) Raw images, (b) the guided images, (c) the over-enhanced results by minimizing (2), (d) the results by minimizing (5).

Fig. 5. The 'bell' structure of $G(I_c)$.

To solve the above problem, an extended adaptive color and contrast enhancement (ACCE) framework is subsequently developed, which is mathematically expressed as:

$$
\begin{cases}
E(I) = \dfrac{1}{2} \sum_{\theta \in h,s,i} \sum_{x} \left(Cr_\theta(x) - I_\theta(x)\right)^2 \\
\quad - \dfrac{\alpha}{2} \sum_{x} G\left(I_s(x)\right) H\left(I_s(x)\right) \sum_{y} w(x,y)\left(I_s(x) - I_s(x)\right)^2 \\
\quad - \dfrac{\beta}{2} \sum_{x} G\left(I_i(x)\right) H\left(I_i(x)\right) \sum_{y} w(x,y)\left(I_i(x) - I_i(y)\right)^2 \\
\quad - \dfrac{\alpha}{2} \sum_{x} \left(1 - H\left(I_s(x)\right)\right) \sum_{y} w(x,y)\left(I_s(x) - I_s(y)\right)^2 \\
\quad - \dfrac{\beta}{2} \sum_{x} \left(1 - H\left(I_i(x)\right)\right) \sum_{y} w(x,y)\left(I_i(x) - I_i(y)\right)^2 \\
s.t. \begin{cases} I_s(x) = 0 \quad I_s(x) < 0 \\ I_s(x) = 1 \quad I_s(x) > 1 \end{cases} , \\
s.t. \begin{cases} I_i(x) = 0 \quad I_i(x) < 0 \\ I_i(x) = 1 \quad I_i(x) > 1 \end{cases}
\end{cases}
\tag{5}
$$

To avoid over-flow, we design a Gaussian weight function $G(I_s)$ and $G(I_i)$ to adaptively adjust the weight of the two corresponding terms to achieve better contrast and brightness. The expression of $G(I_c)$ is given by:

$$
G(I_c) = e^{\frac{-\left((I_c) - 0.5\right)^2}{2\sigma}}
\tag{6}
$$

where $\sigma$ is positive parameter that represents a distribution parameter, $c \in \{s, i\}$ denotes the color channel. In fact, $G(I_c)$ is a 'bell' structure, as displayed in Fig. 5. It can be observed from Fig. 5 that the weight distribution shows a higher trend when pixels located with intermediate values. On the contrary, it has a litter effect on the brighter or darker side. According to the distribution, pixels with small or large values approaching 0 or 1 are assigned less weight, while pixels with intermediate values 0.5 gain the highest weights. Therefore, the function $G(I_c)$ can play an important role in preventing some brighter or darker areas from being over-enhanced. Moreover, the variation of $\sigma$ can make the distribution uniform or polarized, thus altering the influence of contrast enhancement in the different regions. This local-condition based strategy enables the over-enhancement and information loss to be manually controlled within an acceptable range.

In (5), the Heaviside function $H(I_c(x))$ is introduced to determine whether the weight function $G(I_c(x))$ works. It is defined as:

$$
\begin{aligned}
H_c(I(x)) &= h(d(x)) \\
&= h\left((I_c(x) - 0.5)\sum_{y} w(x,y)(I_c(x) - I_c(y))\right)
\end{aligned}
\tag{7-a}
$$

$$
h(d(x)) = \begin{cases} 1 & d(x) > 0 \\ 0 & otherwise \end{cases}
\tag{7-b}
$$

Here, $\sum_{y} w(x,y)(I_c(x) - I_c(y))$ denotes a predicted direction of enhancement, which is derived from the derivatives of the regularized term $\sum_{x} \sum_{y} w(x,y)(I_c(x) - I_c(y))^2$ in (5). Combing with (5), (6) and (7), ACCE method will perform different strategy to generate a desired outcome based on different value of $H(I_c(x))$. In one case ( $H(I_c(x))=1$ ), if an input pixel value $I_c(x)$ in (0.5, 1] is increased by the regularized terms (or $I_c(x)$ in [0, 0.5] is reduced), (5) can be transformed into:

$$
\begin{aligned}
E(I) &= \dfrac{1}{2} \sum_{\theta \in h,s,i} \sum_{x} \left(Cr_\theta(x) - I_\theta(x)\right)^2 \\
&\quad - \dfrac{\alpha}{2} \sum_{x} G\left(I_s(x)\right) \sum_{y} w(x,y)\left(I_s(x) - I_s(y)\right)^2 \\
&\quad - \dfrac{\beta}{2} \sum_{x} G\left(I_i(x)\right) \sum_{y} w(x,y)\left(I_i(x) - I_i(y)\right)^2
\end{aligned}
\tag{8}
$$

Accordingly, in another case ( $H(I_c(x))=0$ ), (5) will turn to:

$$E(I) = \frac{1}{2} \sum_{\theta \in h,s,i} \sum_{x} \left(Cr_{\theta}(x) - I_{\theta}(x)\right)^2$$
$$-\frac{\alpha}{2} \sum_{x} \sum_{y} w(x,y)\left(I_s(x) - I_s(y)\right)^2 \qquad (9)$$
$$-\frac{\beta}{2} \sum_{x} \sum_{y} w(x,y)\left(I_i(x) - I_i(y)\right)^2$$

which is actually consistent with (2).

To implement the minimization of energy in (5), we employ a gradient descent strategy to solve it. In fact, the problem of (5) can be decomposed into:

$$E_1(I_h) = \frac{1}{2} \sum_{x} \left(Cr_h(x) - I_h(x)\right)^2 \qquad (10\text{-a})$$

$$E_2(I_s) = \frac{1}{2} \sum_{x} \left(Cr_s(x) - I_s(x)\right)^2$$
$$-\frac{\alpha}{2} \sum_{x} G(I_s(x)) H(I_s(x)) \sum_{y} w(x,y)\left(I_s(x) - I_s(y)\right)^2 \quad (10\text{-b})$$
$$-\frac{\alpha}{2} \sum_{x} \left(1 - H(I_s(x))\right) \sum_{y} w(x,y)\left(I_s(x) - I_s(y)\right)^2$$

$$E_3(I_i) = \frac{1}{2} \sum_{x} \left(Cr_i(x) - I_i(x)\right)^2$$
$$-\frac{\beta}{2} \sum_{x} G(I_i(x)) H(I_i(x)) \sum_{y} w(x,y)\left(I_i(x) - I_i(y)\right)^2 \quad (10\text{-c})$$
$$-\frac{\beta}{2} \sum_{x} \left(1 - H(I_i(x))\right) \sum_{y} w(x,y)\left(I_i(x) - I_i(y)\right)^2$$

To apply the descent strategy, we firstly need to compute the Euler-Lagrange equation of (10). However, it is inefficient to directly calculate the regularized terms, since the join of $H(I_c(x))$ make the original structure more complex. To solve it, following the strategy in [51], [52], we apply an approximated methodology that reckon it as a constant $H_c(x)$, whose value is directly computed by using (7) in each iteration. Thus, the Euler-Lagrange equation of (10) can be accordingly modified as:

$$\nabla E_1(I_h) = \left(Cr_h(x) - I_h(x)\right) = 0 \qquad (11\text{-a})$$

$$\begin{cases} \nabla E_2(I_s) = \\ \left(I_s(x) - Cr_s(x)\right) - \alpha Do_s(I_s(x)) - \alpha Di_s(I_s(x)) = 0 \\ with \\ Do_s(I_s(x)) \\ = \sum_{y} \left(G(I_s(y)) H_s(y) + G(I_s(x)) H_s(x)\right) w(x,y)\left(I_s(x) - I_s(y)\right) \\ -\frac{I_s(x) - 0.5}{2\sigma} G(I_s(x)) H_s(x) \sum_{y} w(x,y)\left(I_s(x) - I_s(y)\right)^2 \\ and \\ Di_s(I_s(x)) = \sum_{y} w(x,y)\left(I_s(x) - I_s(y)\right)\left(2 - H_s(x) - H_s(y)\right) \end{cases}$$
$$(11\text{-b})$$

**Algo.1 The computation of operator $D$**

1. Input parameter: image $I$, window radius $w$.
2. Output parameter: $D$

3. **function Speed_Up ($I$, $w$)**
4.     Subsample $I$ to construct the k-level Gaussian pyramid $L$, with Size($L[k]$) $\leq w$, where $L[k]$ the kth level.
5.     **for** $j$ from $k$-1 to 1 **do**
6.       **if** Size($L[j+1]$) $\leq w$ **do**
7.         $D[j+1] \leftarrow$ Zeros(Size($L[j+1]$));
8.         **break if**;
9.       **else**
10.         $D[j+1] \leftarrow$ Resize($D[j+1]$, Size($L[j]$));
11.         $L[j+1] \leftarrow$ Resize($L[j+1]$, Size($L[j]$));
12.         $P[j] \leftarrow$ Compute_D($L[j]$, $w$);
13.         $P[j+1] \leftarrow$ Compute_D($L[j+1]$, $w$);
14.         $a \leftarrow sqrt(mse(L[j]))$;
15.         $b \leftarrow sqrt(mse(L[j+1]))$ $st.$ $b \in [0.2*a, 0.8*a]$;
16.         Normalize the coefficients $a$, $b$;
17.         $D[j] \leftarrow D[j+1] + a*P[j] - b*P[j+1]$;
18.       **end if**
19.     **end for**
20.     **return** $D[1]$;
21. **end function**

22. **function Compute_D($L$, $w$)**
23.     Calculate the operator $D$ for each pixel by a sliding window in $L$, with radius set as $w$;
24.     **return** the computational result;
25. **end function**

$$\begin{cases} \nabla E_3(I_i) = \\ \left(I_i(x) - Cr_i(x)\right) - \beta Do_i(I_i(x)) - \beta Di_i(I_i(x)) = 0 \\ with \\ Do_i(I_i(x)) = \\ \sum_{y} \left(G(I_i(y)) H_i(y) + G(I_i(x)) H_i(x)\right) w(x,y)\left(I_i(x) - I_i(y)\right) \\ -\frac{I_i(x) - 0.5}{2\sigma} G(I_i(x)) H_i(x) \sum_{y} w(x,y)\left(I_i(x) - I_i(y)\right)^2 \\ and \\ Di_i(I_i(x)) = \sum_{y} w(x,y)\left(I_i(x) - I_i(y)\right)\left(2 - H_i(x) - H_i(y)\right) \end{cases}$$
$$(11\text{-c})$$

Next, we apply the gradient descent strategy, the evolution follows $\frac{\delta I}{\delta t} = -\nabla E(I)$, where $t$ represents the timeline. For (5), after an explicit discretion with $t$, we can obtain:

$$I_h^{k+1} = \left(1 - \Delta t\right) I_h^k + \Delta t \cdot Cr_h \qquad (12\text{-a})$$

$$\begin{cases} I_s^{k+1} = (1-\Delta t) I_s^k + \Delta t \cdot Cr_s + \alpha \Delta t \cdot Do_s^k + \alpha \Delta t \cdot Di_s^k \\ s.t. \begin{cases} I_s(x) = 0 & I_s(x) < 0 \\ I_s(x) = 1 & I_s(x) > 1 \end{cases} \end{cases} \quad \text{(12-b)}$$

$$\begin{cases} I_i^{k+1} = (1-\Delta t) I_i^k + \Delta t \cdot Cr_i + \beta \Delta t \cdot Do_i^k + \beta \Delta t \cdot Di_i^k \\ s.t. \begin{cases} I_i(x) = 0 & I_i(x) < 0 \\ I_i(x) = 1 & I_i(x) > 1 \end{cases} \end{cases} \quad \text{(12-c)}$$

During the solutions, the computation of operator of $Do$ and $Di$ is inestimable. Because for each pixel, we need to calculate it with all the other pixels in the image to obtain the computational results. To accelerate this progress, we apply a speed-up strategy based on the Gaussian pyramid to solve it. Firstly, we roughly estimate the global information on the low-resolution map of pyramid, and then update it to obtain an accurate one layer by layer. In this manner, we narrow the calculation scope from the whole image to the adjacent region, which can reduce the computational complexity from $O(N^4)$ to $O(N^2 \log N)$. For clarity, the solutions of operator of $Do$ and $Di$ are summarized in Algo.1. Moreover, to hold the convergence of (5), we introduce the following stopping criteria for terminating iterations:

$$\left| \frac{E(I^{k+1}) - E(I^k)}{E(I^k)} \right| \le \tau \quad (13)$$

where $E$ is the energy function, $\tau$ is an extremely small value.

## IV. EXPERIMENTAL RESULTS

In this section, we first discuss the parameter sensitivity of the proposed ACCE method, as well as its performance on denoising effect. Then, we compare our proposed ACCE-D method with state-of-the-art methods qualitatively and quantitatively to demonstrate its superiority. Finally, we also extend the proposed ACCE-D method to other degradation scenarios. All the experiments are implemented in the Matlab R2016b on a Windows 10 platform with 3.6 GHz CPU and 8GB RAM.

### A. Performance of proposed method

In this subsection, in order to examine the effectiveness of the proposed method, we describe and analyze some experimental results in respect to the influence of parameters and denoising effect.

#### 1) Parameter sensitivity

There are three parameters (i.e., $\alpha$, $\beta$, $\sigma$) in our functional energy, as shown in (5) and (6). Following, we study the influence of these three parameters.

We first analyze the effect of parameter $\alpha$. $\alpha$ is the coefficient of saturation terms, which controls the enhancement degree on the S channel in HSI color space. In the test, we implement the proposed method by setting $\alpha \in \{0.05, 0.15, 0.25, 0.4\}$ and fixing $\beta=0.3$, $\sigma=0.03$. It can be observed from Fig. 6(a)-(d) that the saturation in some regions

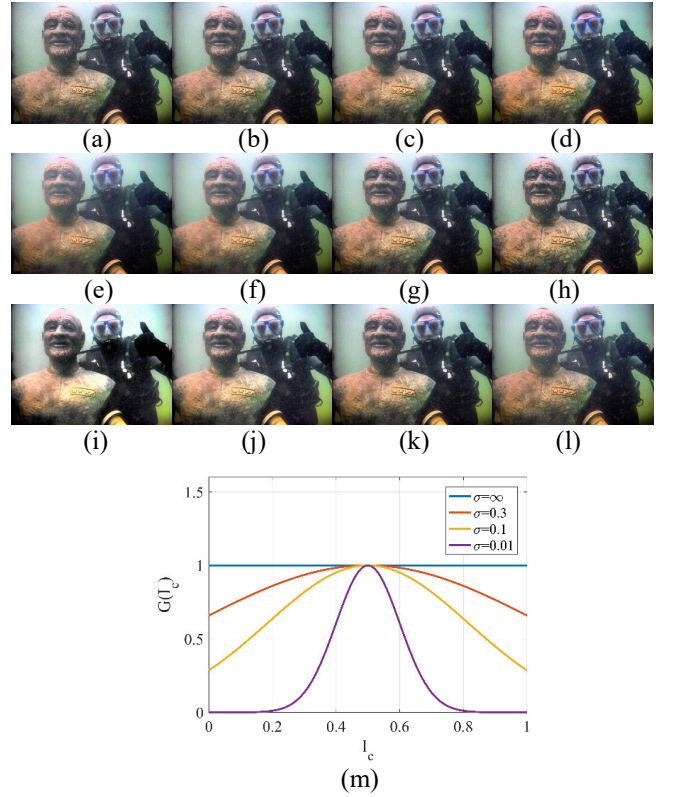

(a) (b) (c) (d)
(e) (f) (g) (h)
(i) (j) (k) (l)

(m)

Fig. 6. The results of varying different parameters. (a-d) the results with configurations: $\beta=0.3$, $\sigma=0.03$ and $\alpha=0.05, 0.15, 0.25, 0.4$, respectively; (e-h) the results with configurations: $\alpha=0.25$, $\sigma=0.03$ and $\beta=0.05, 0.15, 0.3, 0.4$, respectively; (i-l) the results with configurations: $\alpha=0.25$, $\beta=0.3$ and $\sigma=\infty, 0.3, 0.1, 0.01$, respectively; (m) the corresponding curves of $G(I_c)$ with $\sigma=\infty, 0.3, 0.1, 0.01$.

of the enhanced version is increased with the raising of $\alpha$. We can see in Fig. 6(a) that if $\alpha$ is too small (i.e., $\alpha=0.05$), although the result performers well in terms of contrast, showing a low-saturation appearance. On the contrary, when $\alpha$ is too large (i.e., $\alpha=0.4$), the color of the enhanced result seems unnatural since the saturation is over enhanced. In general, we find that $\alpha \in [0.15, 0.3]$ is suitable. More specifically, $\alpha=0.25$ is an appropriate configuration, resulting in a more perceptual color performance.

Similar to $\alpha$, $\beta$ is the coefficient of intensity terms. To test its effect, we vary $\beta$ in the set $\{0.05, 0.15, 0.3, 0.4\}$, and fix other two parameters on the baseline configurations (i.e., $\alpha=0.25$, $\sigma=0.03$). In Fig. 6(e)-(h), we can see that a smooth variation of contrast appears in the enhanced result as $\beta$ increases. In Fig. 6(a) and Fig. 6(b), we see that when $\beta \le 0.15$, the results are not enhanced significantly, leading to some residual haze in scene. This problem is subsequently solved by taking a lager $\beta$, as shown in the Fig. 6(g), where the haze appearance is effectively removed. However, it may causally bring about the noisy result. To balance the contrast, noise and natural appearance, we find that $\beta \in [0.15, 0.35]$ is suitable. In the following experiments, we set $\beta$=0.3 . A lot of experiments have shown that varying $\alpha$ and $\beta$ may lead to different results. However, the other results corresponding to the other values of $\alpha$ and $\beta$ are still acceptable, which demonstrate that our ACCE method is more robust to $\alpha$ and $\beta$ .

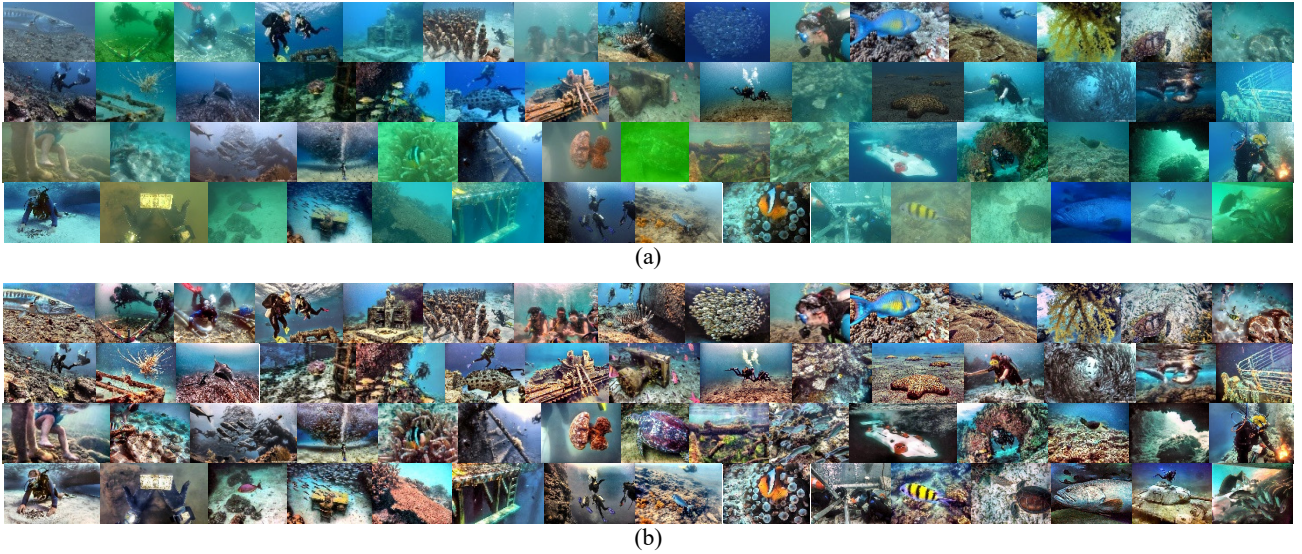

Fig. 7. Performance of the proposed method. (a) Raw underwater images, (b) the corresponding enhanced results by using ACCE-D method.

Next, we study the quality of the enhanced results for different $\sigma$ . As discussed in (6), $\sigma$ indicates a distribution parameter of the proposed Gaussian weight. As a special case when $\sigma \rightarrow \infty$ , the weight becomes a constant 1, and no longer play a role. Here, the proposed approach is executed again by fixing the $\alpha$=0.25 and $\beta$=0.3 , and $\sigma$ is varied in the set $\{\infty, 0.3, 0.1, 0.01\}$ . The enhanced results are displayed in Fig. 6(i)-(l) accompanied with the corresponding curves of proposed Gaussian weight, as show in Fig. 6(m), where we can appreciate how the reduction of $\sigma$ produces a suppression in the over-enhanced regions. Note that vary small $\sigma$ will also influence a global effect of the proposed framework, as reflected in the Fig. 6(l). In our experiments, we find that the ACCE method presents a better result when $\sigma$ is set to be 0.03.

For better viewing the results with declared configurations, i.e., $\alpha$=0.25, $\beta$=0.3, $\sigma$=0.03, $\Delta$t=0.7, $\tau$=0.05 , we provide 60 raw underwater images captured under various conditions for test, as shown in Fig. 7(a). Their corresponding enhanced versions are accordingly presented in Fig. 7(b). In Fig 7, one can easily see the quality of the raw images is significantly improved. For these underwater images with different degraded types, the proposed method is able to recovery the abundant color dynamic as well as high visibility. Even for some extreme deteriorated scenarios, ACCE-D method is still effective in the color and contrast enhancement, which preliminarily proved the superiority of our scheme.

### 2) Evaluation on noise reduction

In this subsection, we further carry out some tests to evaluate the improvement of ACCE in suppressing noise by introducing the high-frequency denoising scheme. Fig. 8 illustrates two examples to demonstrate the excellent performance in denoising by using the proposed scheme. As shown in Fig. 8(b), the cropped red region appears more undesired noise when compared with the original underwater image in Fig. 8(a). But comparing with Fig. 8(c), we can observe that our noise reduction strategy can effectively reduce the noise and preserve more edges and details. For more objective evaluation, we further exploit two commonly used measures PSNR and SSIM [53] to assess the performance of proposed ACCE-D method. Since there is no available ground truth database for underwater images, we calculate the PSNR and SSIM values between the original underwater images and the corresponding recovered ones. Likewise, the higher PSNR and SSIM scores in Fig. 8(c) also demonstrate that our schemes can effectively avoid the noise amplification, as well as improve the accuracy of enhanced results.

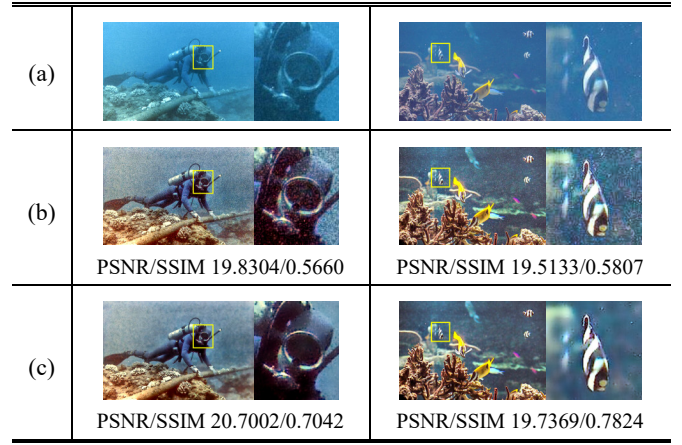

Fig. 8. Examples of the denoising effect. (a) Raw images, (b) the results without using the high-frequency denoising strategy, (c) the results by using our ACCE-D method.

### B. Qualitative comparison

To demonstrate the superiority of proposed method, in this part we compare our results versus several state-of-the-art methods including ACE [47], RCP [17], IBLA [20], UMSR [33], L²UWE [30] and GNCE [40]. We conduct some experiments on six types of underwater images, which are captured under the greenish, blueish, blue-greenish, hazy, low-light and turbid scenes, respectively, as shown in Fig. 9(a). Their enhanced and restored version are accordingly displayed in Fig. 9(b)-(h). In Fig. 9(b), it can be seen that ACE method can effectively enhance the contrast and thus bring an improvement in visibility. But it fails to achieve visual appealing results due to the inaccurate color correction. RCP and IBLA methods restore the degraded image based on the physical model, which seem to have less effect on seriously distorted image. More specifically, for the scene with insufficient illumination, RCP method even reduces the contrast of dark regions, resulting the poor visibility, as presented in the Fig. 9(c). In spite of the well performance in edge sharpness, IBLA method is prone to over enhance the input image, especially for the image with uneven lighting, as shown in Fig. 9(d). On the contrary, we can see from Fig. 9(f) that UMSR method unveils more details in dark region by making illumination uniform. However, it introduces considerable artifacts and unnatural grey appearance. Likewise, as an enhancer specific for low-light image, the L²UWE method focuses on the detail and edge, while rarely considers the color deviation. Last but not least, The GNCE method can alleviate the problem of image color deviation to some extent, but it is not good at improving the contrast. Compare with these methods, our ACCE-D method has the best visual quality, since it not only recovery the genius color, but also significantly improve the clarity of scenes. Besides, unlike the others contrast enhancement algorithms, our approach does not oversaturate pixel values so that the details are better retained or even highlighted. We can better view this argument via a careful inspection on the seaweed in the low-light scene in Fig. 9(h).

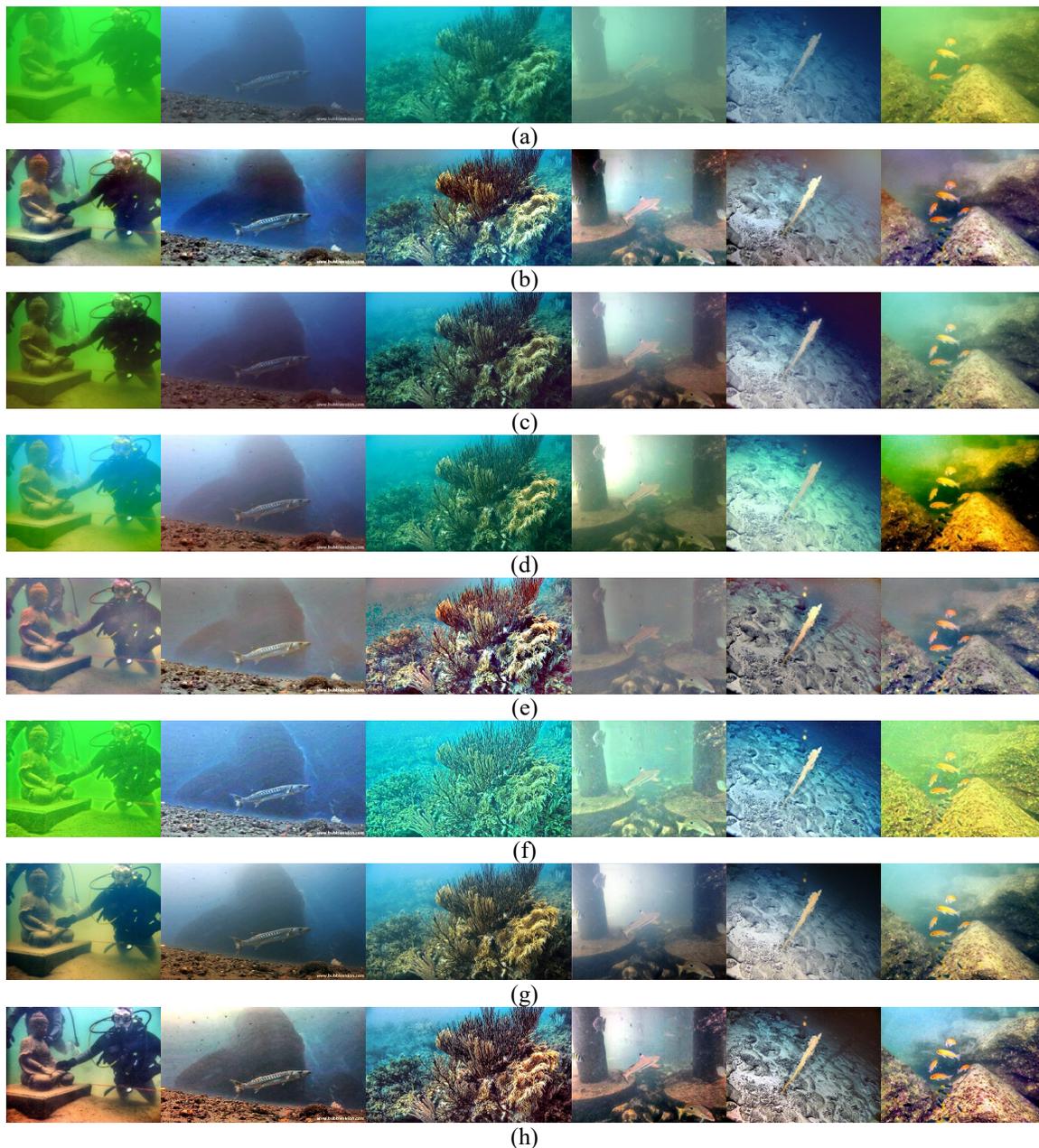

Fig. 9. Comparisons of different methods on various challenging underwater scenes. (a) Raw images captured under different challenging scenes, (b)-(h) the enhanced and restored results by using ACE, RCP, IBLA, UMSR, L²UW, GNCE and the proposed method, respectively.

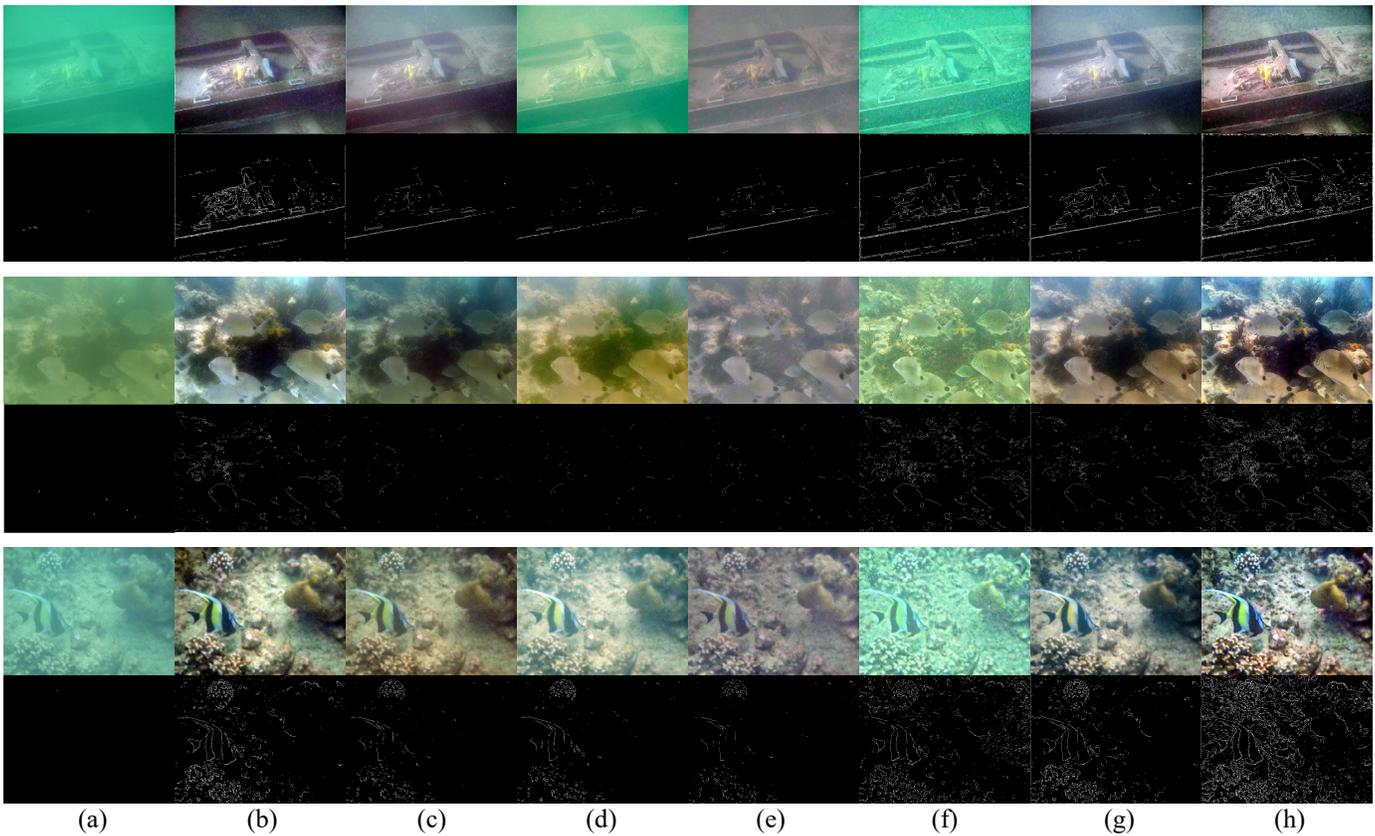

Fig. 10. Comparisons of the Canny edge results. (a) Raw images and their visible edge maps from canny operator; (b-h) the corresponding results of ACE, RCP, IBLA, UMSR, L²UWE, GNCE and proposed method, respectively.

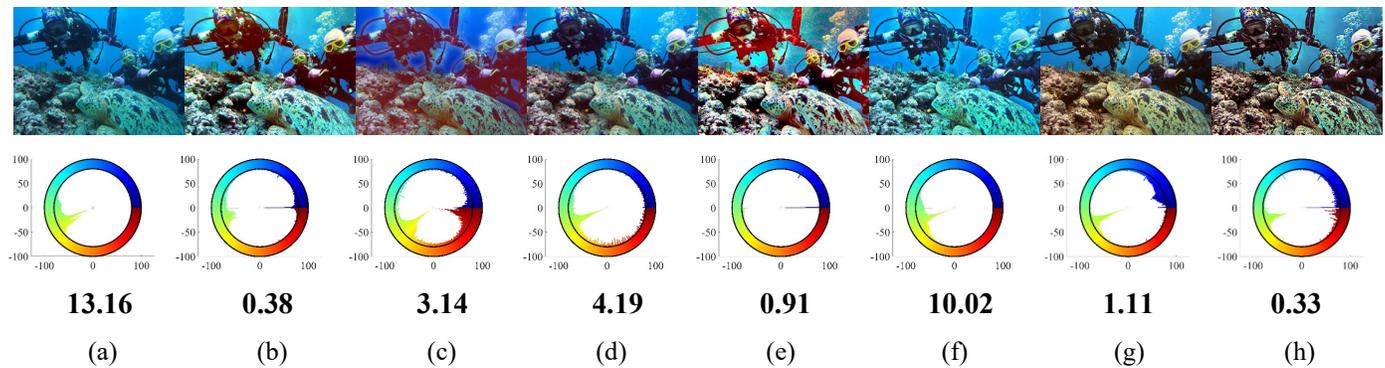

| **13.16** | **0.38** | **3.14** | **4.19** | **0.91** | **10.02** | **1.11** | **0.33** |
| (a) | (b) | (c) | (d) | (e) | (f) | (g) | (h) |

Fig. 11. Comparisons of color performance. (a) Raw image and its polar hue histograms, (b-h) the corresponding results of ACE, RCP, IBLA, UMSR, L²UWE, GNCE and proposed method, respectively. The score of color deviation detection [55] is also presented on the bottom of each image.

To assess the amount of clarity improved, we depend on the Canny edge detection [54] for purpose of visible edge recovery evaluation on images with serious deterioration in visibility. Fig. 10 presents the enhanced and restored results and their corresponding edge maps. One can easily observed that the edges of original images can hardly be detected due to the intense scattering. For these challenging scenes, RCP, IBLA, UMSR methods seem to leave fog in the results, as reflected on the edge map. Compared with the original image, the retrieved edge does not increase much. On the contrary, a larger number of visible edges is reproduced by our methodology. Observing from Fig. 10(g), almost all the contours of our enhanced images are contained in the detected results, which indicates that the proposed ACCE-D can effectively remove the influence of foggy appearance as well as reveal more details of image structure.

For the color correction, since there is no ground-truth image to refer to, we utilize the polar hue histograms to assess it. Besides, a metric of color deviation detection [55] is also adopted to rate the results of each algorithm. The wider the diverse chromaticity of histogram and the smaller the score of detection represent a more vivid color performance. In order to better compare their difference, we choose a typical blueish image to test, as shown in Fig. 11. At first glance on the hue histogram, the result of RCP method appears to have the best chromatic diversity. However unfortunately, its seemingly good distribution of red hues is actually due to RCP method over enhancing the red channel. It is proved by the careful inspection on the restored result of RCP, wherein the black area is reddish. In all other case, the chromaticity of our result performs more diversity than the compared state-of-the-art methods, which indicates that the proposed method has the highest accuracy for color correction. This is also demonstrated by the subsequent detection of color cast. In test, the best score is gained by our result, and is far less than the original image. Therefore, we can conclude that the proposed approach subjectively outperforms other methods in terms of contrast and color accuracy.

Table I Quantitative comparisons of UCIQE (The best result is in bold).

| | Greenish | Blueish | Blue-greenish | Hazy | Low-light | Turbid | Average |
|---|---|---|---|---|---|---|---|
| Original | 0.3928 | 0.4801 | 0.5170 | 0.3930 | 0.5579 | 0.4893 | 0.4717 |
| ACE | 0.6111 | 0.5833 | 0.6464 | 0.6058 | 0.5732 | 0.6085 | 0.6047 |
| RCP | 0.5866 | 0.5549 | 0.6095 | 0.5869 | 0.5652 | 0.5677 | 0.5785 |
| IBLA | 0.5781 | 0.5999 | 0.5838 | 0.5366 | 0.5639 | **0.6916** | 0.5923 |
| UMSR | 0.5056 | 0.5154 | 0.6562 | 0.3677 | 0.5548 | 0.5504 | 0.5250 |
| L$^2$UWE | 0.4118 | 0.5148 | 0.5380 | 0.4302 | 0.6151 | 0.5075 | 0.5029 |
| GNCE | 0.6446 | 0.6133 | 0.6449 | 0.5878 | 0.5901 | 0.6203 | 0.6168 |
| New | **0.6785** | **0.6385** | **0.6917** | **0.6118** | **0.6215** | 0.6561 | **0.6462** |

Table II Quantitative comparisons of UIQM (The best result is in bold).

| | Greenish | Blueish | Blue-greenish | Hazy | Low-light | Turbid | Average |
|---|---|---|---|---|---|---|---|
| Original | 0.6035 | 0.6397 | 1.1978 | 0.4064 | 1.1935 | 0.6840 | 0.7875 |
| ACE | 1.4767 | 1.2209 | 1.5114 | 1.0466 | 1.5112 | 1.1195 | 1.3144 |
| RCP | 1.1203 | 0.8347 | 1.4650 | 0.8309 | 1.3750 | 0.8531 | 1.0798 |
| IBLA | 1.0727 | 0.9817 | 1.3538 | 0.7902 | 1.3954 | 1.2984 | 1.1487 |
| UMSR | 1.1720 | 0.9359 | 1.4767 | 0.6344 | 1.7375 | 1.0671 | 1.1706 |
| L$^2$UWE | 1.3464 | 1.0479 | 1.4943 | 0.9546 | 1.6756 | 1.1044 | 1.2705 |
| GNCE | 1.2492 | 0.9563 | 1.4188 | 0.8844 | 1.4615 | 1.0172 | 1.1646 |
| Ours | **1.6090** | **1.2621** | **1.6603** | **1.2263** | **1.8889** | **1.3542** | **1.5001** |

Table III Quantitative comparisons of PCQI (The best result is in bold).

| | Greenish | Blueish | Blue-greenish | Hazy | Low-light | Turbid | Average |
|---|---|---|---|---|---|---|---|
| Original | 1.0000 | 1.0000 | 1.0000 | 1.0000 | 1.0000 | 1.0000 | 1.0000 |
| ACE | 1.2661 | 1.0747 | 1.1859 | 1.1364 | 1.0628 | 1.0352 | 1.1269 |
| RCP | 1.0112 | 0.9471 | 1.0824 | 1.0471 | 1.0301 | 0.9555 | 1.0122 |
| IBLA | 1.0956 | 1.0191 | 1.0683 | 1.0384 | 1.0105 | 1.0890 | 1.0535 |
| UMSR | 1.2358 | 1.0232 | 1.2260 | 1.0286 | 1.2277 | 1.1878 | 1.1549 |
| L$^2$UWE | 1.3083 | 1.1037 | 1.2611 | 1.1460 | 1.2906 | 1.1460 | 1.2232 |
| GNCE | 1.1290 | 1.0437 | 1.1800 | 1.0545 | 1.1009 | 1.0726 | 1.0968 |
| Ours | **1.3167** | **1.1228** | **1.3555** | **1.1878** | **1.3940** | **1.2683** | **1.2743** |

Table IV Quantitative comparisons of CPBD (The best result is in bold).

| | Greenish | Blueish | Blue-greenish | Hazy | Low-light | Turbid | Average |
|---|---|---|---|---|---|---|---|
| Original | 0.5858 | 0.3074 | 0.3467 | 0.5506 | 0.7108 | 0.3426 | 0.4740 |
| ACE | 0.8398 | 0.6953 | 0.4047 | 0.6383 | 0.7169 | 0.4138 | 0.6181 |
| RCP | 0.7575 | 0.3360 | 0.3422 | 0.5671 | 0.6809 | 0.3497 | 0.5056 |
| IBLA | 0.8332 | 0.4860 | 0.3482 | 0.6138 | 0.6941 | 0.4807 | 0.5760 |
| UMSR | 0.8218 | 0.5329 | 0.3974 | 0.5755 | 0.7116 | 0.3799 | 0.5699 |
| L$^2$UWE | 0.8307 | **0.7670** | 0.3627 | 0.6547 | 0.6989 | 0.5092 | 0.6372 |
| GNCE | 0.8038 | 0.5382 | 0.3538 | 0.5587 | 0.7079 | **0.5587** | 0.5869 |
| Ours | **0.8483** | 0.7582 | **0.4661** | **0.6680** | **0.7445** | 0.4782 | **0.6590** |

Table V Quantitative comparisons of average values of UCIQE, UIQM, PCQI, CPBD of 200 random selected raw images (The best result is in bold).

| Metric | ACE | RCP | IBLA | UMSR | L$^2$UWE | GNCE | Ours |
|---|---|---|---|---|---|---|---|
| UCIQE | 0.6107 | 0.5576 | 0.5398 | 0.5620 | 0.5102 | 0.5975 | **0.6454** |
| UIQM | 1.4772 | 1.2760 | 1.3223 | 1.4105 | 1.4471 | 1.3825 | **1.6035** |
| PCQI | 1.2005 | 1.0621 | 1.1279 | 1.2372 | 1.2994 | 1.1815 | **1.3642** |
| CPBD | 0.6043 | 0.6079 | 0.6199 | 0.5993 | 0.6007 | 0.5985 | **0.6478** |

### C. Quantitative comparison

To objectively test the proposed method, we exploit four widely used metrics including UCIQE [56], UIQM [57], CPBD [58] and PCQI [59]. CBPD calculates the accumulative probability of detecting blur to score the image. PCQI is a patch-based index that focus on the image contrast variation. In contrast to CPBD and PCQI, UIQM and UCIQE are more comprehensive indicators specifically designed for evaluating underwater image quality. UCIQE is a linear combination of chroma, saturation and contrast to quantify the image quality. Similarly, UIQM scores the outcomes based on the colorfulness, sharpness and contrast. For all the metrics, the higher score indicates a better result.

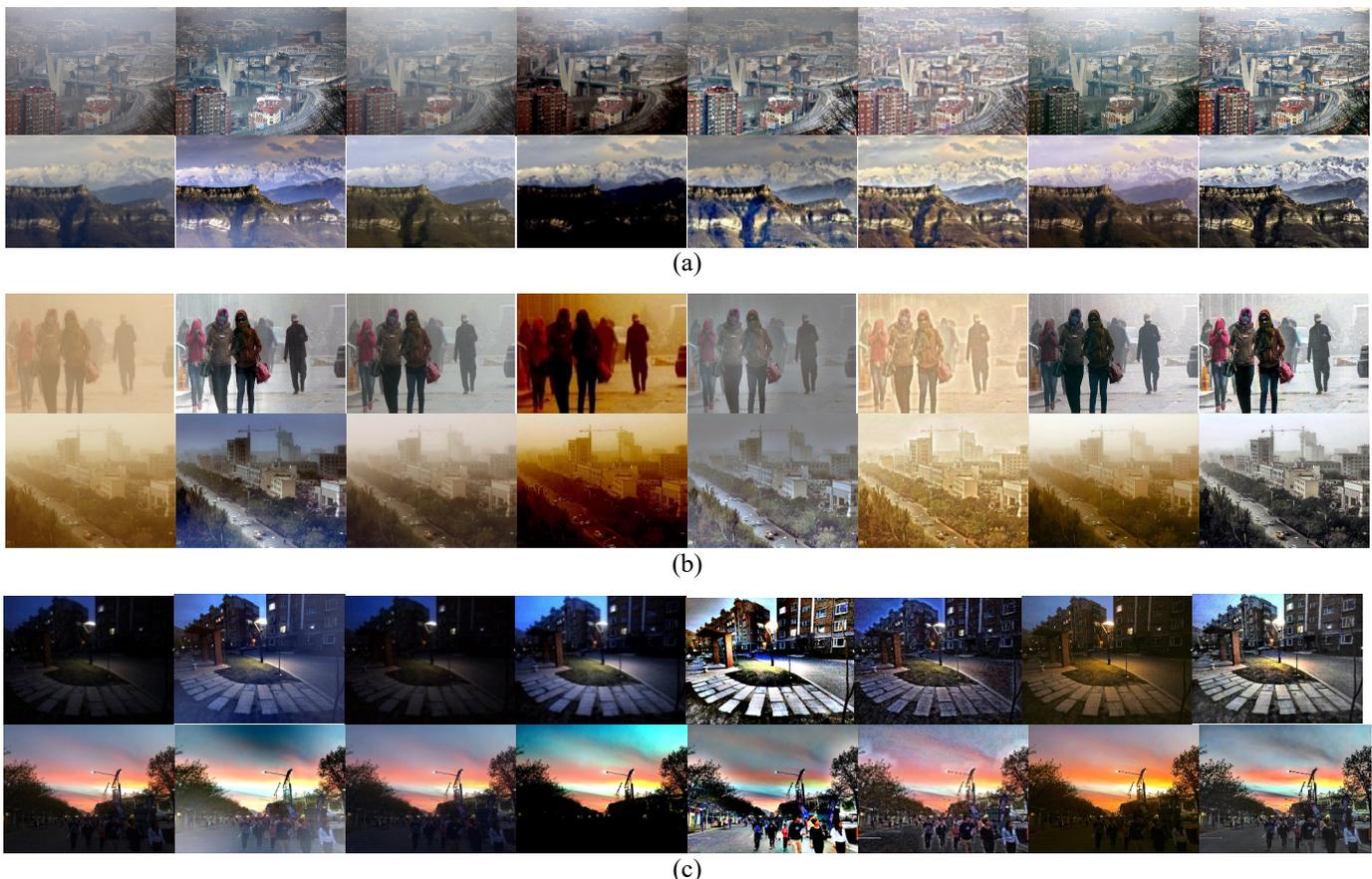

Fig. 12. Comparisons on other types of degraded images: (a) foggy images, (b) sandstorm images and (c) low-illumination images. From left to right: raw images and the enhanced and restored results of ACE, RCP, IBLA, UMSR, L²UWE, GNCE and proposed method, respectively.

The assessment results associated to Fig. 9 are displayed in Tables I-V. The best score of each column has been marked in bold. It can be observed that our method outperforms others in most cases. For different kinds of degraded images, the UCIQE scores of our results is above 0.6, and UIQM is also higher than 1.2. When move on to CBPD and PCQI, our method still received the highest evaluation due to the more stable and outstanding scores. The best UCIQE, UIQM, CBPD and PCQI indicates that the proposed method can significantly improves the contrast and visibility of degraded image, meanwhile recovers more vivid color, which is consistent with our subjective analysis. To make our experimental results more convincing, we conduct a broader test using 200 randomly selected degraded images from underwater benchmark dataset [42]. The average values of UCIQE, UIQM, CBPD, and PCQI of these five compared methods and our ACCE-D method are counted and presented in Table V. Again, the optimal scores reveal that ACCE-D achieves more robustness than the other techniques. We have hence demonstrated the superiority of the proposed method.

### D. Other applications

Despite our method is designed for underwater environment, it can be generalized to other types of degraded images, due to it doesn't rely on any model assumptions or physical processes. As mentioned before, the atmospheric foggy images have the similar imaging formation model to that of underwater scenes, which can be adopted to test their ability in terms of descattering. Fig. 12(a) presents the results of different methods on atmospheric foggy images. We observe that the ACE and GNCE methods produce unexpected color deviation in the results. For the model-based methods, RCP fails to remove the hazy layer, while IBLA reduces contrast and brightness of dark regions. In the results of UMSR, some residual hazy can be also found in remote objects. Intuitively, L²UWE method can effectively increase the visible edges and details. But unfortunately, it seems to retain the haze color in the enhanced results. On the contrary, the best dehazing effect is achieved by the ACCE-D due to the outperforms local and global contrast in outcomes. Besides, our results are more natural since their color performance are close to the subjective perception.

Compared with the foggy images, the sandstorm images contain more serious yellowish color cast. In such cases, for each algorithm, there are still the problems that have appeared in the foggy scenes, as shown in Fig. 12(b). Although RCP and UMSR methods can correct the color bias in some degree, they are unavailable to eliminate the scattered effect. IBLA method further aggravates the color cast. Visually, the outcomes generated by the proposed scheme show more genius color and higher clarity, which demonstrate its effectiveness.

The low light images are characterized by the deficient ambient light. For these degraded images, the model-based methods usually erroneously estimate the depth, and thus fail to work. In Fig. 12(c), we can easily find that RCP and GNCE method produce inappreciable influence in illumination recovery. IBLA method provides unstable outputs because it may produce some invisible dark areas, as shown in the fourth column of Fig. 12(c). UMSR is more aggressive in raising the image exposure, which is beneficial to unveil more details. Nevertheless, some over-enhanced areas and amplified noise can be found in enhanced results. Likewise, ACE and L$^2$UWE methods introduce the artifacts into their outcomes. In contrast, our ACCE-D can well balance the noise and illumination, as well as slightly improve the color performance. Therefore, the superiority and versatility of ACCE-D are further verified.

## V. CONCLUSION

In this paper, we have presented a novel framework for enhancing underwater image via adaptive color and contrast enhancement and denoising, in which we decompose the image into high- and low-frequency components by using the DoG filter and bilateral filter, respectively. For the high-frequency part, a simple soft-thresholding algorithm is adopted to reduce the noise and preserve the edges. In low-frequency component, we design a variational framework for adaptively enhancing the contrast and color, which can significantly improve the visibility and highlight the details of scene without excessive enhancement. Numerous experiments on real-word underwater image with different challenge scenes demonstrate that the proposed method is robust and effective in contrast enhancement, color correction and noise suppression. The extensively qualitative and quantitative experimental results further validate its superiority comparing with other advanced methods. In addition, the application tests verified the proposed ACCE-D can be extend to enhance other kind of degraded scenarios.